\documentclass[review]{elsarticle}

\usepackage{lineno,hyperref}
\usepackage{graphicx}

\usepackage{amsmath,amssymb}
\usepackage{mathrsfs}
\usepackage{color}
\usepackage{graphics}
\usepackage{multirow}
\usepackage{caption}
\usepackage{subcaption}
\usepackage{amssymb}
\usepackage{array}
\usepackage{amsmath,bm}
\usepackage{colortbl}
\usepackage{dsfont}
\usepackage{diagbox}
\usepackage{rotating}
\usepackage{booktabs}
\usepackage{overpic}
\usepackage{contour}
\usepackage{bbding}
\usepackage[marginal]{footmisc}
\usepackage{pifont}
\usepackage{pifont}
\usepackage{booktabs}
\usepackage{makecell}
\usepackage{verbatim}
\usepackage{bbding}
\usepackage{url}

\usepackage{algorithm}
\usepackage{algorithmic}

\newcommand{\tabincell}[2]{\begin{tabular}{@{}#1@{}}#2\end{tabular}}
\newcommand{\figref}[1]{Fig.~\ref{#1}}
\newcommand{\tabref}[1]{Table~\ref{#1}}

\newcommand{\secref}[1]{$\S$ \ref{#1}}
\definecolor{mygray}{gray}{.92}
\definecolor{mypink}{RGB}{251, 81, 169}
\newcommand{\eg}{\emph{e.g.}}
\newcommand{\ie}{\emph{i.e.}}
\newcommand{\etal}{\emph{et al.}}

\newcommand{\etc}{\emph{etc}}

\newcommand{\rev}[1]{{\textcolor{black}{#1}}}

\newcommand{\ourmodel}{\textit{ERRNet}}
\graphicspath{{./Imgs/}}
\DeclareGraphicsExtensions{.jpg,.pdf,.png}

\modulolinenumbers[5]

\journal{Journal of \LaTeX\ Templates}

\begin{document}

\begin{frontmatter}

\title{Fast Camouflaged Object Detection via \\Edge-based Reversible Re-calibration Network}

\author[whu]{Ge-Peng Ji}
\author[hkust-gz]{Lei Zhu}
\author[cug]{Mingchen Zhuge}
\author[scu]{Keren Fu\corref{mycorrespondingauthor}}\cortext[mycorrespondingauthor]{Corresponding author}\ead{fkrsuper@scu.edu.cn}
\address[whu]{School of Computer Science, Wuhan University}
\address[hkust-gz]{Hong Kong University of Science and Technology (Guangzhou)}
\address[cug]{School of Computer of Science, China University of Geosciences, China}
\address[scu]{College of Computer Science, Sichuan University}

\begin{abstract}
Camouflaged Object Detection (COD) aims to detect objects with similar patterns (\eg, texture, intensity, colour, \etc) to their surroundings, and recently has attracted growing research interest.
As camouflaged objects often present very ambiguous boundaries, how to determine object locations as well as their weak boundaries is challenging and also the key to this task.
Inspired by the biological visual perception process when a human observer discovers camouflaged objects, this paper proposes a novel edge-based reversible re-calibration network called \ourmodel.
Our model is characterized by two innovative designs, namely Selective Edge Aggregation (\textbf{SEA}) and Reversible Re-calibration Unit (\textbf{RRU}), which aim to model the visual perception behaviour and achieve effective edge prior and cross-comparison between potential camouflaged regions and background. More importantly, RRU incorporates diverse priors with more comprehensive information comparing to existing COD models.
Experimental results show that \ourmodel~outperforms existing cutting-edge baselines on three COD datasets and five medical image segmentation datasets.
Especially, compared with the existing top-1 model SINet, \ourmodel~significantly improves the performance by $\sim$6\% (mean E-measure) with notably high speed (79.3~FPS), showing that \ourmodel~could be a general and robust solution for the COD task.
\end{abstract}


\end{frontmatter}


\section{Introduction}

Camouflaged object detection (COD) aims to detect concealed/camouflaged objects\footnote{Following the same definition of COD in~\cite{fan2020Camouflage}, we mainly focus on finding animals/people from the natural scene rather than artificial patterns.}
in visual scenes.
Such objects are grand-master of camouflaging or concealing themselves in their environment with the capability to mimic the same body colours, patterns, as well as other morphological appearance of the background, \eg, dazzle camouflage~\cite{scott2011dazzle}, mimesis~\cite{meyer2006repeating}, disruptive coloration~\cite{stevens2006disruptive}, and distractive markings~\cite{dimitrova2009concealed}.
COD techniques can be adopted in numerous potential applications, including agriculture (\eg, disaster detection~\cite{amit2016analysis}, locust detection~\cite{yi2019locust,piotr2020mobile}), underwater image processing (\eg, object detection/segmentation~\cite{rizzini2015investigation}), computer-assisted search and rescue~\cite{fan2020Camouflage}, and medical imaging (\eg, lung infection diagnosis~\cite{fan2020InfNet,wu2021jcs}, retinal image segmentation~\cite{zhang2019attention}, and polyp detection~\cite{fan2020pra,ji2021pnsnet}).

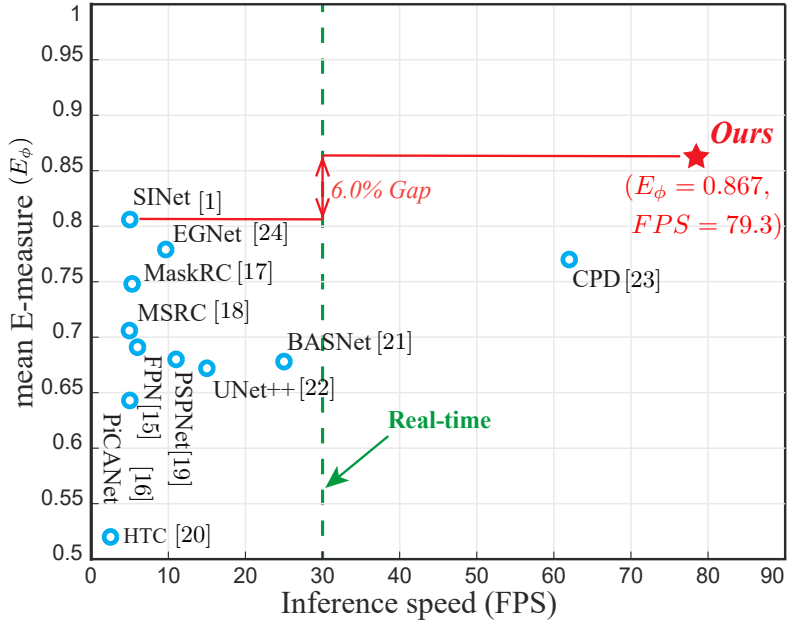
\begin{figure}[t!]
\centering
\begin{overpic}[width=0.85\linewidth]{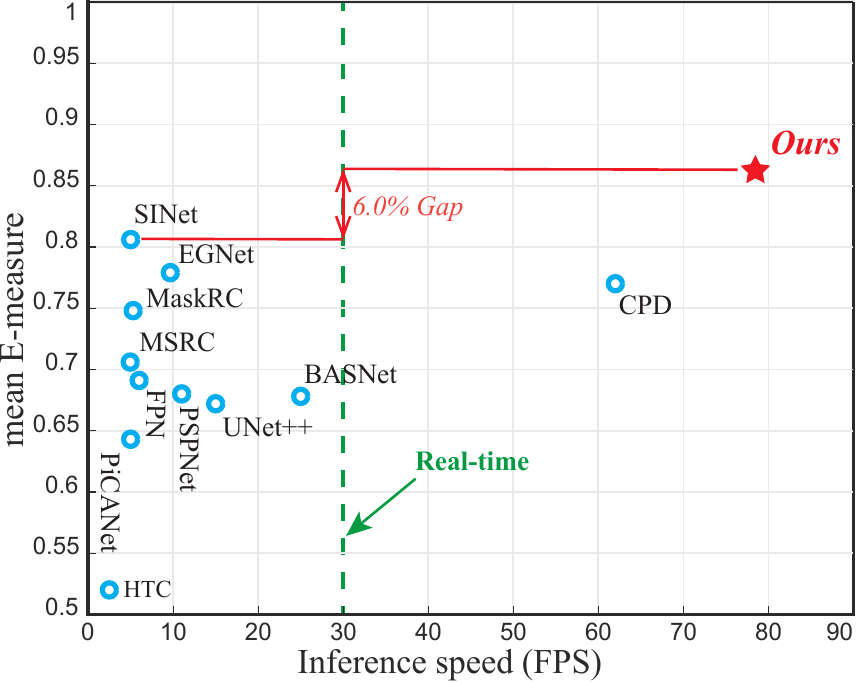}
	\put(79, 55){\textcolor{red}{($E_\phi = 0.867$,}}
	\put(80, 50){\textcolor{red}{$FPS=79.3$)}}
	\put(2, 56){\begin{rotate}{90}\small ($E_\phi$) \end{rotate}}
	\put(17, 28){\begin{rotate}{-90}\small \cite{lin2017feature}\end{rotate}}
	\put(16, 20){\begin{rotate}{-90}\small \cite{liu2018picanet}\end{rotate}}
	\put(29, 44){\small \cite{he2017mask}}
	\put(26, 39){\small \cite{huang2019mask}}
	\put(21, 22){\begin{rotate}{-90}\small \cite{zhao2017pyramid}\end{rotate}}
	\put(21, 10){\small \cite{chen2019hybrid}}
	\put(47, 35){\small \cite{qin2019basnet}}
	\put(37, 29){\small \cite{zhou2019unet++}}
	\put(79, 43){\small \cite{wu2019cascaded}}
	\put(31, 49){\small \cite{zhao2019EGNet}}
	\put(24, 53){\small \cite{fan2020Camouflage}}
\end{overpic}
\caption{\small Inference Speed (\ie, FPS) \textit{v.s.}~mean E-measure ($E_\phi$) on COD10K~\cite{fan2020Camouflage} dataset. The proposed~\ourmodel~achieves competitive performance and faster inference speed compared to the SOTA camouflaged object detection methods.}
\label{fig:inference_speed}
\end{figure}

\rev{However, well addressing COD is very challenging due to the following reasons:
(i) High variation in object sizes.
(ii) Ambiguous object boundaries.
(iii) Texture similarity between the objects and their environments. 
Thus, as mentioned by Fan~\etal~\cite{fan2020Camouflage}, ``\textit{The high intrinsic similarities between the target object and the background make COD far more challenging than the traditional object detection task.}''.
As a closely related task, we note that salient object detection (SOD) aims to identify the most attention-grabbing regions/objects in visual scenes.
In general, the term ``salient'' exists essentially orthogonality relation to ``camouflaged'', \ie, conspicuous objects vs. concealed objects.
Providing a comprehensive review of the SOD task is beyond the scope of this paper. 
We refer readers to recent surveys~\cite{Fan2021SOC,wang2021salient} and technical approaches~\cite{zhuge2021salient,fu2019deepside,Siris_2021_ICCV} for more details.
}
%

Biological vision studies~\cite{merilaita2017camouflage} have shown that, when finding camouflaged objects from the current visual scene (``the background'') whose patterns are almost identical to those of objects, an observer will initially pop out possible region proposals (which we call \textit{\textbf{global prior}});
then, an observer will shift his/her attention~\cite{wang2020salient,zhang2020attention,qin2020u2} to the surroundings from these potential regions.
Some of these unreasonable regions, as a result, will be excluded individually via \textit{\textbf{comparing}} them and their surroundings; 
and during the comparison, the holistic contour of the target is depicted in virtue of the supplement of detailed cues (which we call \textit{\textbf{edge prior}});
finally, the camouflaged objects can be identified correctly.
According to the above hypotheses, the following three conclusions are achieved: (i) Weak boundary cues are the key to the COD task; (ii) Except for the boundary cues, the global guidance regions are essential for proposing potential camouflaged regions; (iii) The cross-comparison stage of human visual perception for comparing between potential targets and their surroundings should be explicitly modelled. 
Unfortunately, existing COD models~\cite{fan2020Camouflage,le2019anabranch} seldom completely consider the above biologically inspired process, therefore leading to less satisfactory results.

%

%
To this end, we propose a novel framework, named~\ourmodel,~for more accurately detecting camouflaged objects. \ourmodel~is highlighted by two innovative designs, namely Selective Edge Aggregation (\textbf{SEA}) and Reversible Re-calibration Unit (\textbf{RRU}). SEA and RRU aim to simulate the biological visual perception process, achieving effective edge prior and cross-comparison between camouflaged objects and background. As shown in \figref{fig:inference_speed}, it achieves a competitive 0.867 E-measure, while maintaining a very efficient inference speed 79.3~FPS when comparing to several cutting-edge approaches. It is worth noting that the top-1 model SINet only achieves 5~FPS with $6\%$ lower accuracy.
The main \textbf{contributions} of this paper are three-fold:

\begin{itemize}

    \item We propose a novel architecture, called Edge-based Reversible Re-calibration Network (\textbf{\ourmodel}), for camouflaged object detection, which acts as a strong baseline to spark novel ideas on this task. The proposed~\ourmodel~greatly advances the state-of-the-art (SOTA) performance with real-time inference speed on eight challenging datasets.

    \item We design an aggregation strategy, Selective Edge Aggregation (\textbf{SEA}), to obtain initial edge prior, which can well alleviate the ``ambiguous'' problem of weak boundaries.

    \item To model the cross-comparison stage of visual perception, we further design a multivariate calibration strategy, termed Reversible Re-calibration Unit (\textbf{RRU}), which re-calibrates the coarse inference map by considering \textbf{NEGS} priors (\ie,~Neighbour prior, Global prior, Edge prior, and Semantic prior) to contrast potential camouflaged regions and their complement areas. This is able to compensate the previously mentioned global prior and edge prior with more comprehensive information. 


\end{itemize}


\section{Related Work}\label{sec:related}

In this section, we provide detailed reviews on camouflaged object detection (COD) methods.
Aside from this, we discuss edge-based convolutional neural networks (CNNs) models in many binary segmentation tasks and highlight the difference between these methods and our edge-based COD network.

\noindent
\textbf{Camouflaged Object Detectors.} \ Early methods detected camouflaged objects by designing various hand-engineered features, which evaluate the similarities between camouflaged regions and the background details.
Pan~\etal~\cite{pan2011study} studied a 3D convexity based method to identify camouflaged objects in images.
Liu~\etal~\cite{liu2012foreground} presented a foreground object detection scheme by integrating the top-down information with the bottom-up cues based on the
expectation maximisation framework.
Sengottuvelan~\etal~\cite{sengottuvelan2008performance} used a co-occurrence matrix method to detect camouflaged objects from a single input image.
Yin~\etal~\cite{hou2011detection} proposed an optical flow model to realise the detection of the mobile object with camouflage colours.
Gallego~\etal~\cite{gallego2014foreground} employed a region-based spatial-colour Gaussian Mixture Model (GMM) to model the foreground object for segmenting the foreground object in moving camera sequences.
These hand-crafted features based methods work well for only a few cases where images/videos have a simple and non-uniform background.
However, the camouflage detection performance degrades largely when there is a
strong similarity between the foreground and the background.

Motivated by the dominated performance of CNNs in diverse computer vision tasks, a few researchers developed CNNs to address the problem of detecting camouflaged objects on a large number of annotated data.
Le~\etal~\cite{le2019anabranch} built an image dataset (\ie, CAMO) of camouflaged objects, and developed an end-to-end network with a classification stream and a segmentation stream for camouflaged object segmentation.
The classification branch predicted the probability of containing camouflaged objects in the input image and then fused the probability map into the segmentation branch for improving COD segmentation accuracy.
Fan~\etal~\cite{fan2020Camouflage} developed a network (SINet) with two major modules, \ie, the search module (SM) and the identification module (IM).
The SM is responsible for searching for a camouflaged object while IM is then used to
precisely detect it.
\rev{There are also some concurrent works published after the submission of this paper.
For examples, Lyu~\etal~\cite{Lyu2021Mutual} propose the first joint SOD and COD network within an adversarial learning framework to explicitly model prediction uncertainty~\cite{zhang2021uncertainty} of each task.
Mei~\etal~\cite{mei2021Ming} develop a novel distraction mining strategy for distraction discovery and removal, which is adopted to build a positioning and focus network.
Fan~\etal~\cite{fan2021concealed} develop an enhanced model upon \cite{fan2020Camouflage} to achieve promising performance, where two well-elaborated sub-components are proposed, including neighbour connection decoder and group-reversal attention.
}
Yang~\etal~\cite{yang2021uncertainty} propose a framework using a probabilistic representational model combined with transformers to explicitly reason uncertainties in the camouflaged scene.

\noindent
\textbf{Edge-aware Binary Image Segmentation.} \
Recently, learning additional edge information has shown superior performance in many binary image segmentation tasks.
A few salient object detection (SOD) works~\cite{luonon2017,li2018contour,liu2019simple,wu2019cascaded,feng2019attentive,qin2019basnet,zhang2019capsal,wu2019mutual} learned edge cues to assist the saliency inference.
Luo~\etal~\cite{luonon2017} proposed a U-shape model with an IoU edge loss to directly optimise edges of target saliency maps.
Feng~\etal~\cite{feng2019attentive} proposed a boundary-aware loss between the predicted and annotated saliency maps. 
Qin~\etal~\cite{qin2019basnet} implicitly injected the goal of accurate boundary prediction in a hybrid loss, which combined a binary cross-entropy (BCE), structural
similarity (SSIM), and IoU losses. 
Liu~\etal~\cite{liu2019simple} designed multiple pooling-based modules and built an edge detection branch to further sharpen the details of salient objects.
Wang~\etal~\cite{wang2019salient} designed a salient edge detection module to emphasise the importance of salient edge information.
Zhao~\etal~\cite{zhao2019EGNet} designed a network to leverage the complementarity between salient edge information and salient object information.
Wu~\etal~\cite{wu2019stacked} stacked cross refinement units into a CNN to refine the multi-level features of salient object detection and edge detection.
Apart from SOD, Chen~\etal~\cite{chen20MTMT} explored shadow edges, shadow count, and shadow regions and unlabelled data for shadow detection. A detailed review of the SOD model is out of the scope of this work, the reader can refer to the recently released benchmark paper~\cite{fan2018salient} for more details.

Different from the above works, we propose a new scheme to achieve more accurate localisation and boundary delineation of camouflaged objects. The obvious differences and advantages of the proposed scheme are that we incorporate diversified priors in the decoder stage for more promising results, and we also show that these priors promote the performance for COD. 

\begin{figure*}
\includegraphics[width=0.99\textwidth]{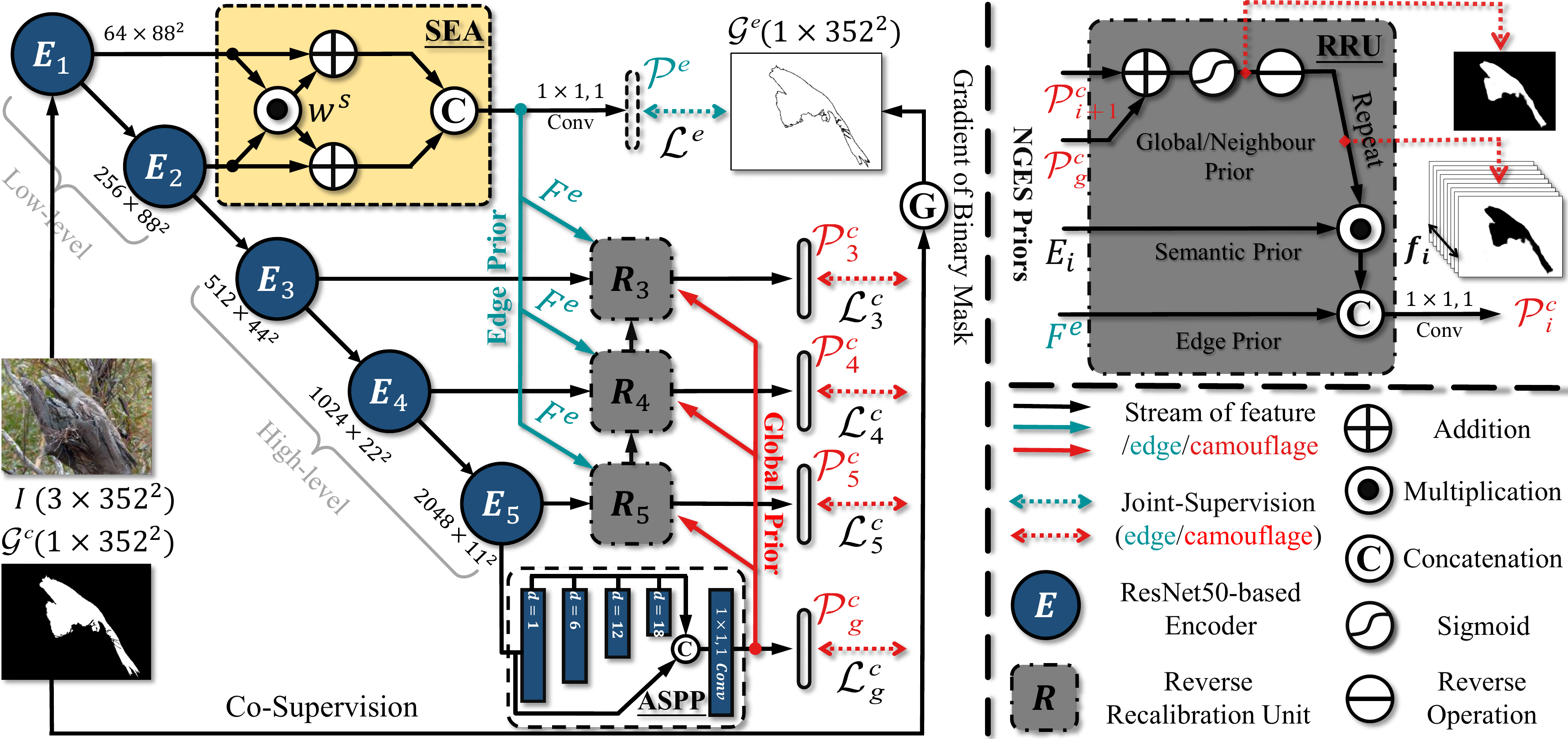}
\caption{\small
The overall pipeline of the proposed \ourmodel~that contains three main cooperative components, including Atrous Spatial Pyramid Pooling (ASPP) for initiating global prior, Selective Edge Aggregation (SEA) for generating edge prior, and Reversible Re-calibration Unit (RRU) for modulating and refining the NGES Priors in a cascaded manner. More details are described in~\secref{sec:problem_formulation}.}
\label{fig:architecture}
\end{figure*}


\section{Proposed Framework}

\subsection{Problem Formulation}\label{sec:problem_formulation}

\begin{figure*}[t]
\includegraphics[width=0.99\textwidth]{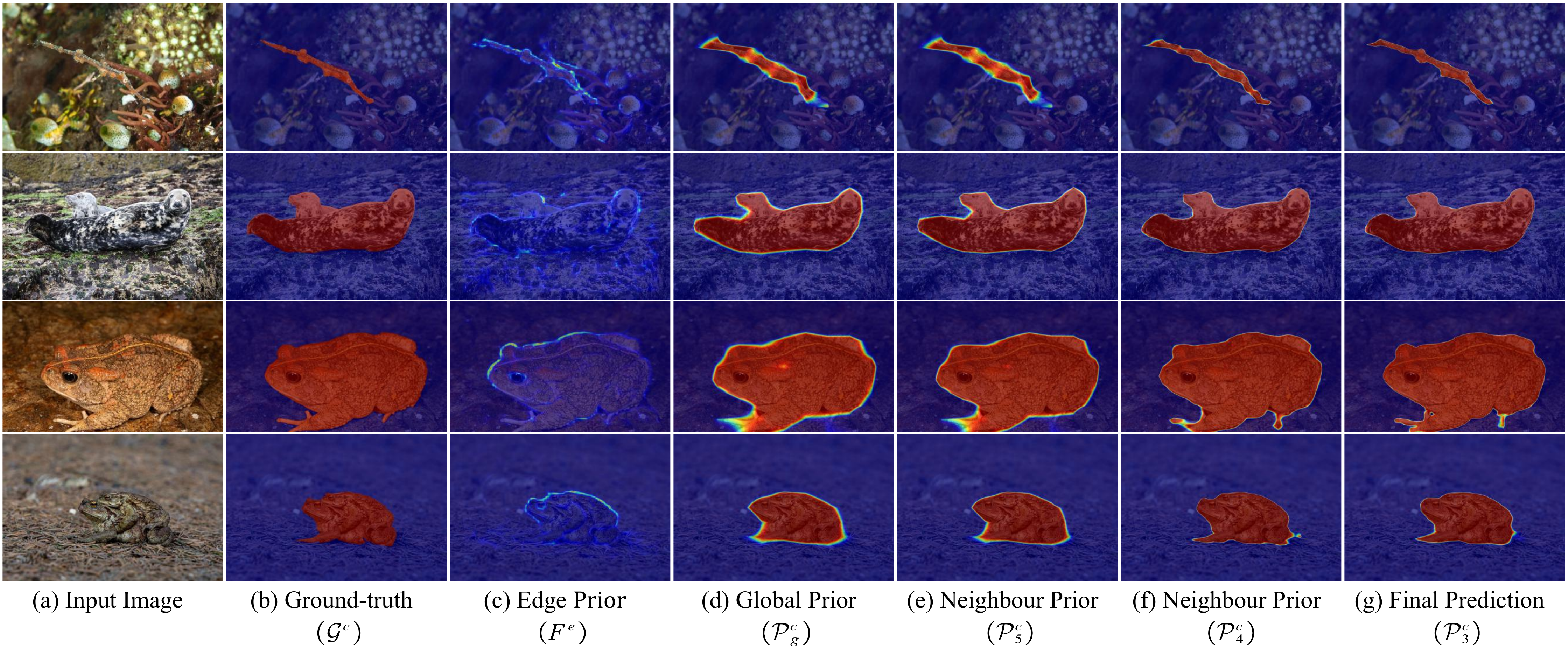}
\caption{
\small Visualization of each component in the NEGS priors, \ie,~edge prior in (c), global prior in (d), and neighbour prior in (e) \& (f). Specifically, the re-calibration stage treats the intermediate outputs of the network as the prior cues to enhance the reliability and stability of the learning process, and thus, more accurate final prediction (g) is obtained. Note that since the semantic prior $E_i$ is directly borrowed from the features of the ResNet-50 backbone, it is not shown here.
}
\label{fig:feature_visualization}
\end{figure*}

The overall framework of the proposed network, named~\ourmodel, is shown in~\figref{fig:architecture}. Given an input image $\mathcal{I} \in \mathds{R}^{H \times W \times 3}$ that contains at least one camouflaged object in the scene,
we first utilise ResNet-50 backbone~\cite{he2016deep} (after removing the fully-connected layers) to abstract five feature hierarchies from low-level ($E_1~\&~E_2$) to high-level ($E_3,~E_4,~\&~E_5$), which shares a similar spirit to~\cite{wu2019cascaded}. As stated in~\cite{chen2017deeplab}, cascaded atrous convolution module (Atrous Spatial Pyramid Pooling, ASPP) contributes to enlarging receptive filed and can fuse multi-context information via adopting four different dilation rates ($d \in \{1,6,12,18\}$), and thus, we employ it in the last feature hierarchy of encoder ($E_5$) to acquire the global prior $\mathcal{P}^{c}_{g}$. The formulation can be defined as:
\begin{equation}
    \mathcal{P}^c_g = \mathcal{F} \left(Cat \left(E_5,~\mathcal{F}^A_1(E_5),~\mathcal{F}^A_{6}(E_5),~\mathcal{F}^A_{12}(E_5),~\mathcal{F}^A_{18}(E_5)\right)\right),
\end{equation}
in which $Cat(\cdot)$ and $\mathcal{F}(\cdot)$ denote concatenation operation and convolution block followed by a 1$\times$1 filter, respectively. Symbol $\mathcal{F}^A_d$ denote atrous convolution block with dilation rate $d$. 

Then, we design a novel aggregation strategy (\textbf{S}elective \textbf{E}dge \textbf{A}ggregation, \textbf{SEA}) to selectively fuse the low-level edge-aware features $\{E_1~\&~E_2\}$. Additionally, we constraint it with the explicit edge supervision $\mathcal{L}^e$, which is conducive to providing valuable edge prior in the re-calibration phase. Details about edge aggregation strategy are given in~\secref{sec:SEA}.

To enhance the reliability of the prediction, we introduce a multivariate calibration strategy (\textbf{R}eversible \textbf{R}e-calibration \textbf{U}nit, \textbf{RRU}) to reversely re-calibrate the coarse prediction via combining NGES priors to contrast potential camouflaged regions and their complement areas (indicated by the reversed object maps). It can be viewed as the progressive refinement with diversified priors in an cascaded fashion. NEGS priors consists of four components, namely the neighbour prior \textbf{map} $\{\mathcal{P}^{c}_{i+1},~i \in \{3,4,5\}\}$, global prior \textbf{map} $\mathcal{P}^{c}_{g}$, edge prior \textbf{features} $F^e$, and semantic prior \textbf{features} $\{E_i,~i \in \{3,4,5\}\}$. In \figref{fig:feature_visualization}, we provide the intuitive visualisation map of each prior. Detailed formulations of RRU are described in~\secref{sec:RRU} and Algorithm~\ref{alg:RRU}.

Finally, the network produces high-quality binary prediction masks $\{\mathcal{P}^c_i,~i \in \{3,4,5,g\}\}$ after Sigmoid operation to mark concealed objects, in which each pixel lies in $[0,1]$, indicating the probability of being the camouflaged object. Specifically, $\mathcal{P}^c_i=1$ indicates that a pixel is fully identified as belonging to the camouflaged object, and $\mathcal{P}^c_i=0$ means the opposite. Details are in \secref{sec:loss_function}.

\vspace{-5pt}
\subsection{Selective Edge Aggregation}\label{sec:SEA}

To provide more discriminative representation in the features, we utilise explicit edge information as auxiliary cues in the decoder stream. These well-endowed low-level representations serve as complementary information in the re-calibration stream, which can facilitate the RRU module to depict more fine-grained contour of camouflaged objects.
Generally, both visual cues and noises such as boundary, corner, line, and irrelevant high-frequency signal, \etc, are preserved by low-level layers. Further, in the high-level layer, the down-sampling operation in CNNs may cause coarse vision perception. Therefore, directly fusing low-level features into high-level ones may result in redundancy and inconsistency.

To address the aforementioned problems, we first design a novel Selective Edge Aggregation (SEA) module that adaptively picks up mutual representations from inputs before aggregation.
As shown in~\figref{fig:architecture}, SEA consists of two branches with the inputs ($E_1 \in \mathds{R}^{\frac{H}{4} \times \frac{W}{4} \times 64}$ and $~E_2 \in \mathds{R}^{\frac{H}{4} \times \frac{W}{4} \times 256}$) received from the encoder, and then, the convolution block $ \mathcal{F}_{i}^{e}(\cdot) $ with 64 filters is adopted to reduce the channel numbers,
which can be formulated as follows:
\begin{equation}
    \bar{E}_{1} = \mathcal{F}_{1}^{e} (E_1),~\bar{E}_{2} = \mathcal{F}_{2}^{e} (E_2).
\end{equation}

Then, we design a selective weighted switcher $w^s$ to adaptively learn how to weigh different attention on different levels ($i=1,2$). The process can be defined as element-wise multiplication:
\begin{equation}
    w^s = \mathcal{F}(\bar{E}_{1} \odot \bar{E}_{2}),
\end{equation}
which can be conceptualized as the `linear attention'. It mainly contributes to suppressing background noise and putting more attention on interesting region via weighting each other, however, this procedure may cause  valuable cues vanishing problem simultaneously. Therefore, we design a residual aggregation process to fuse the two features:
\begin{equation}
    F^e = \mathcal{F} \left(Cat\left( \mathcal{F}(\bar{E}_{1} \oplus w^s),~\mathcal{F} (\bar{E}_{2} \oplus w^s) \right) \right),
\end{equation}
in which $\mathcal{F}(\cdot)$ and $Cat(\cdot)$ denotes the convolution operation with 64 filters and concatenation operation, respectively.
%
%
It well alleviates the aforementioned problem profiting from the proprieties of residual learning~\cite{he2016deep}. Then, we get the edge prior $F^e$ with 64 channels, which will be exploited in the reversible re-calibration phase.

Additionally, we utilise a convolution block with 1 filter to generate the coarse camouflaged edge $\mathcal{P}^e$ that is supervised by camouflaged edge ground truth $\mathcal{G}^e$. Intuitively, it contributes to learning more edge-aware knowledge with explicit constraints.
These properties enable the network to augment the characteristic details of the features and maintain the consistency of semantic knowledge between different levels.
Visual explanations for the effectiveness of SEA refer to~\figref{fig:module_validation} (c) and (d).

\renewcommand{\algorithmicrequire}{\textbf{Input:}}
\renewcommand{\algorithmicensure}{\textbf{Output:}}
\begin{algorithm}[t]
	\caption{: Reversible Re-calibration Unit (RRU)}
	\label{alg:RRU}
	\begin{algorithmic}[1]
	\REQUIRE{Multi-level features from encoder: $\{E_i,~i \in \{1,2,3,4,5\}\}$}
	\ENSURE{Camouflaged mask prediction: $\{\mathcal{P}^{c}_i,~i \in \{3,4,5\}\}$}
	\STATE{Generate an edge prior: $F^e = SEA(E_1,~E_2)$}
    \STATE{Generate a global prior: $\mathcal{P}^{c}_{g} = ASPP(E_5)$}
	\FOR{$i=5$; $i>2$; $i--$ }
	    \STATE{$//$Reversing gaze and visual focus of attention.}
	    \IF{$i==5$}
	    \STATE{$F^{R}_{i} ~=~ \Psi_{f_i} \left(1 - \mathcal{S}(\mathcal{P}^{c}_{g})\right)$}
	    \ELSE
	    \STATE{$F^{R}_{i} ~=~ \Psi_{f_i} \left(1 - \mathcal{S}(\mathcal{P}^{c}_{i+1} \oplus \mathcal{P}^{c}_{g})\right)$}
	    \ENDIF
	    \STATE{$//$Reversible re-calibration with supplying edge-aware details.}
	    \STATE{$\mathcal{P}^{c}_{i} ~=~ \mathcal{F} \left( Cat(F^e,~F^R_i \odot E_i) \right)$}
	\ENDFOR
	\RETURN{Camouflaged mask prediction: $\{\mathcal{P}^{c}_i,~i  \in \{3,4,5\}\}$}
	\end{algorithmic}
\end{algorithm}

\subsection{Reversible Re-calibration Unit}\label{sec:RRU}
Generally, existing methods only focus on how to stimulate on target regions (positive sides) while overlooking the background regions (negative sides), we design a multivariate reversible re-calibration unit that is devoted to double-checking the positive/negative sides and fine-tuning the semantic prior in a cascaded fashion. Different from~\cite{huang2017semantic,choe2019attention,chen2018reverse}, we further probe the relationship of edge-aware details, foreground, and background for addressing the disorientation from similar patterns between camouflaged foreground (target-class) and non-camouflaged background (reverse-class).
Particularly, we elaborately utilise: (i) the reverse operation to mine the complementary regions and details via erasing the estimated camouflaged regions from high-level side-output semantic priors; and (ii) the supplement of edge prior to locating around the boundaries of camouflaged regions.


\begin{figure}[!t]
\centering
\includegraphics[width=\linewidth]{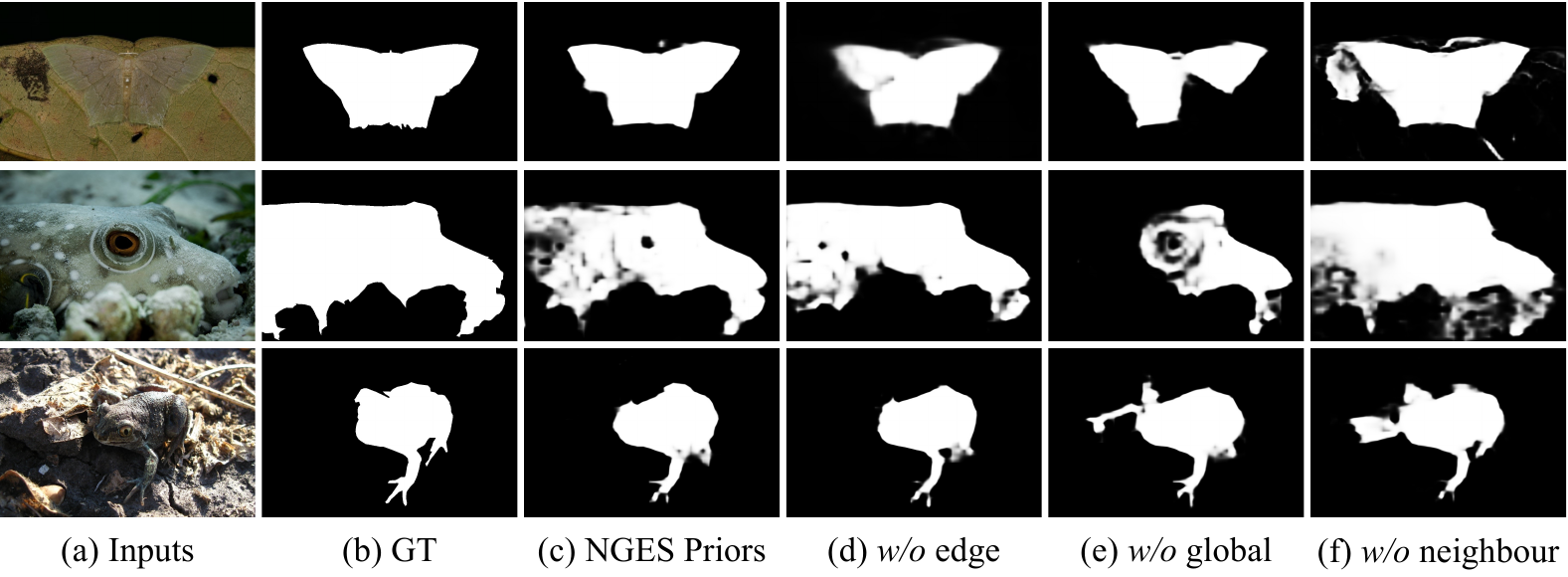}
\caption{\small The effectiveness of NGES priors in RRU. (b) ``GT'' and (c) ``NGES Priors'' means the ground truth and full model of~\ourmodel, respectively. We observe that: (d) ``edge prior'' promotes the fine-grained of ambiguous weak boundaries; (e) ``global prior'' helps locate potential camouflaged regions; and (f) ``neighbour prior'' enhances the stability of prediction when global prior is less satisfactory or unavoidable noise is introduced. Since none of the \textit{NGES prior} is dispensable, they have their own importance.}
\label{fig:module_validation}
\end{figure}

As shown in the right sub-figure of~\figref{fig:architecture}, NGES priors are fed to each RRU $\{R_i,~i \in \{3,4,5\}\}$, including the neighbour prior maps $\{P^{c}_{i+1},~i \in \{3,4,5\}\}$ from higher-level output of RRU ($R_{i+1}$), edge prior features $F^e$ from SEA, global prior map $\mathcal{P}^{c}_{g}$ from ASPP, and semantic prior features $\{E_i,~i \in \{3,4,5\}\}$ from encoder.
We summarise the re-calibration process of RRU into the following two major steps:

\textit{\textbf{(i) Reversing Gaze and Visual Focus of Attention.}}
For the camouflaged priors $\mathcal{P}^{c}_{g}$ \& $\{\mathcal{P}^{c}_i,~i \in \{3,4,5\}\}$, we adopt reverse operation to generate reversal camouflaged priors $F^R_i$, which can be formulated as 5$^{th}$$\sim$9$^{th}$ lines of Algorithm~\ref{alg:RRU}.
Symbol $\mathcal{S}(\cdot)$ indicates normalising the inputs into $[0,1]$ via standard \textit{Sigmoid} function.
Additionally, we apply the repeated stacking operation $\Psi_{f_i}(\cdot)$ for expanding single-channel mask into $f_i$ channels that have the same channel numbers as the feature hierarchy $E_i$. It is noted that we assign the neighbour prior $\mathcal{P}^{c}_{i+1}=0$ when $i=5$. The effectiveness of global and neighbour priors is illustrated in~\figref{fig:module_validation} (e) and (f), respectively.

\textbf{\textit{(ii) Providing Edge-aware Details in Reversible Re-calibration.}} For the edge prior $F^e$ (cyan line in~\figref{fig:architecture}), it serves as the supplementary vision details in the re-calibration process, which are learned from the SEA module within low-level layers. Equipped with the above characteristic features, we get the final prediction $\mathcal{P}^{c}_{i}$ via re-calibrating the feature borrowed from the encoder. It can be formulated as 11$^{th}$ line of Algorithm~\ref{alg:RRU}. The symbols $Cat(\cdot)$ and $\mathcal{F}(\cdot)$ denotes the concatenation operation and convolution block with one filter to achieve map prediction, respectively. Finally, each RRU produces a \textbf{prediction map} for highlighting the camouflaged objects, as shown in \figref{fig:feature_visualization}.

Note that our reversible re-calibration mechanism is different from the recent methods in the usage of multi-level features and refinement mechanism. Specifically, instead of simple fusion in the last stage of the decoder in PoolNet~\cite{liu2019simple} and PFA~\cite{zhao2019pyramid}, we utilise an iterable module RRU to refine the coarse map in a cascaded fashion, which is guided by the edge priors in a bottom-up manner and global priors in a top-down manner simultaneously. Rather than split supervision in AFNet~\cite{feng2019attentive} without further feature fusion, our SEA contributes to selectively acquire edge priors from low-level features in a co-supervised manner, which are fed into RRU to generate the more accurate prediction.

\subsection{Loss Function}\label{sec:loss_function}
Pioneer works~\cite{qin2019basnet,F3Net} have found that combining multiple loss functions with adaptive weights at different levels can promote the performance of the network with better convergence speed. Therefore, we utilise the co-supervision strategy to jointly train the multiple branches, which can acquire knowledge from the source domain of multi-modalities (mask- and edge-based ground truth).
The loss functions in our co-supervision strategy can be formulated as:
$\mathcal{L}_{total} = \mathcal{L}^e + \sum_{i \in \{3,4,5,g\}} \mathcal{L}^c_i$,
which consists of two components: (i) For the supervision of camouflaged mask, we consider the significance of both micro-level and macro-level. Thus, our loss consists of a weighted BCE (Binary Cross-Entropy) and a weighted IoU (Intersection-of-Union) loss $\mathcal{L}^c = \mathcal{L}_{BCE}^{\mathcal{W}} + \mathcal{L}_{IoU}^{\mathcal{W}}$; (ii) With respect to the camouflaged edge supervision, we only adopt micro-level loss $\mathcal{L}^e = \mathcal{L}_{BCE}^{\mathcal{W}}$. That is because of the huge imbalance between positive and negative samples, which may lead to a large deviation of macro-level loss during learning. The formulation of these loss functions $\mathcal{L}_{BCE}^{\mathcal{W}}$ and $\mathcal{L}_{IoU}^{\mathcal{W}}$ are the same as in~\cite{F3Net}, and their capability has been verified in the field of salient object detection.


\section{Experiments}

In this section, we introduce the benchmark datasets and the evaluation metrics, as well as conduct experiments to verify the effectiveness of the proposed network.
\textit{We will release the trained model, our results on COD benchmark datasets, the code, and our results of two applications upon the publication of this work.~\footnote{\url{https://github.com/GewelsJI/ERRNet}}}

\vspace{-5pt}
\subsection{Benchmark Protocols}\label{sec:benchmark_protocols}
There are three COD datasets used in our experiments.
The COD10K~\cite{fan2020Camouflage} is the most challenging dataset, which is the largest publicly available annotated COD dataset with 3,040 training images and 2,026 testing images.
CAMO~\cite{le2019anabranch} is another small-scale COD dataset with 1,000 training images and 250 testing images.
CHAM~\cite{2018Animal} contains 76 images, which are all used for testing.
Follow the most recent COD method~\cite{fan2020Camouflage}, we combine the training sets of COD10K (3,040 images) and CAMO (1,000 images) to build the final training set.
Additionally, we test each COD method on the testing sets of COD10K, CAMO, and the whole images of CHAM for obtaining the results of three benchmark datasets for fair comparisons.


\begin{table*}[t!]
  \centering
  \small
  \renewcommand{\arraystretch}{0.8}
  \renewcommand{\tabcolsep}{2.2pt}
  \caption{\small Quantitative results on different datasets, including CHAM, CAMO, and COD10K. The best scores are highlighted in \textbf{bold}. $E_\phi$ denotes mean E-measure~\cite{fan2018enhanced}.  ``N/A'' means the result is not available. Concerning metrics, symbols ``$\uparrow$''/``$\downarrow$'' mean the higher/lower is better.}
  \label{tab:BenchmarkResults}
  \begin{tabular}{r|cccc|cccc|cccc}
  \hline
  \toprule
    & \multicolumn{4}{c|}{\tabincell{c}{CHAM~\cite{2018Animal}}}
   &\multicolumn{4}{c|}{\tabincell{c}{CAMO~\cite{le2019anabranch}}}
   & \multicolumn{4}{c}{ \tabincell{c}{COD10K~\cite{fan2020Camouflage}}} \\
  \cline{2-13}
   Baselines
   &$S_\alpha\uparrow$      &$E_\phi\uparrow$     &$F_\beta^w\uparrow$      &$M\downarrow$
   &$S_\alpha\uparrow$      &$E_\phi\uparrow$     &$F_\beta^w\uparrow$      &$M\downarrow$
   &$S_\alpha\uparrow$      &$E_\phi\uparrow$     &$F_\beta^w\uparrow$      &$M\downarrow$ \\
   \hline
   \hline
 FPN~\cite{lin2017feature}
         &.794&.783&.590&.075&.684&.677&.483&.131&.697&.691&.411&.075\\
 MaskRC~\cite{he2017mask}
     &.643&.778&.518&.099&.574&.715&.430&.151&.613&.748&.402&.080\\
 PSPNet~\cite{zhao2017pyramid}
     &.773&.758&.555&.085&.663&.659&.455&.139&.678&.680&.377&.080\\
 UNet++~\cite{zhou2019unet++}
     &.695&.762&.501&.094&.599&.653&.392&.149&.623&.672&.350&.086\\
 PiCANet~\cite{liu2018picanet}
     &.769&.749&.536&.085&.609&.584&.356&.156&.649&.643&.322&.090\\
 MSRC~\cite{huang2019mask}
     &.637&.686&.443&.091&.617&.669&.454&.133&.641&.706&.419&.073\\
 BASNet~\cite{qin2019basnet}
     &.687&.721&.474&.118&.618&.661&.413&.159&.634&.678&.365&.105\\ 
 PFANet~\cite{zhao2019pyramid}
     &.679&.648&.378&.144&.659&.622&.391&.172&.636&.618&.286&.128\\
 CPD~\cite{wu2019cascaded}
     &.853&.866&.706&.052&.726&.729&.550&.115&.747&.770&.508&.059\\ 
 HTC~\cite{chen2019hybrid}
     &.517&.489&.204&.129&.476&.442&.174&.172&.548&.520&.221&.088\\
 EGNet~\cite{zhao2019EGNet}
     &.848&.870&.702&.050&.732&.768&.583&.104&.737&.779&.509&.056\\
 A-Net~\cite{le2019anabranch}
     & N/A & N/A & N/A & N/A &.682&.685&.484&.126& N/A & N/A & N/A & N/A\\
 SINet~\cite{fan2020Camouflage}
     &.869&.891&.740&.044
     &.751&.771&.606&.100
     &.771&.806&.551&.051\\
 \hline
 \textbf{\ourmodel}
 &\textbf{.877}&\textbf{.927}&\textbf{.805}&\textbf{.036}
 &\textbf{.761}&\textbf{.817}&\textbf{.660}&\textbf{.088}
 &\textbf{.780}&\textbf{.867}&\textbf{.629}&\textbf{.044}\\
  \bottomrule
  \hline
  \end{tabular}
\end{table*}

\vspace{-5pt}
\subsection{Evaluation Metrics}
In our implementation, we choose $\mathcal{P}^{c}_3$ \textbf{as the final segmentation result} since it better combines NGES priors via the reversible re-calibration process. We adopt four widely-used metrics for quantitative comparisons among different COD methods, including:

\textbf{(i) Structure-measure~\cite{fan2017structure} ($S_{\alpha}$)} computes the structural similarity between $\mathcal{P}^c_3$ and $\mathcal{G}^c$ by taking both region-part and object-part into consideration. Its definition is given by:
\begin{equation}  \label{eq_S_measure}
    S_{\alpha}= \alpha \cdot S_o (\mathcal{P}^{c}_{3},~\mathcal{G}^c) + (1-\alpha) \cdot S_r(\mathcal{P}^c_3,~\mathcal{G}^c) \ ,
\end{equation}
where
$S_o$ and $S_r$ are the object-aware and region-aware structural similarity, respectively.
$\alpha=0.5$ balances these two structural similarities, as indicated in~\cite{fan2017structure}.

\textbf{(ii) Mean Enhanced-measure~\cite{fan2018enhanced} ($E_{\phi}$)} devises an enhanced alignment matrix ($\phi_{FM}$) to capture both image-level statistics and local pixel matching information:
\begin{equation} \label{eq_E_measure}
    E_\phi = \frac{1}{W \times H} \sum_{i=1}^{W} \sum_{j=1}^{H}  \phi_{FM}(i, j).   
\end{equation}

\textbf{(iii) Weighted F-measure~\cite{margolin2014evaluate} ($F_{\beta}^{w}$)} uses weighting functions on the precision and recall errors:
%
\begin{equation}  \label{eq_wF_measure}
    F_{\beta}^{w} = \frac{(1+\beta^2)\times Precision^w \times Recall^w}{\beta^2 \times Precision^w + Recall^w},
\end{equation}
where $Precision^w$ and $Recall^w$ denote the weighted precision and recall.
$\beta^2 = 0.3$; please refer to~\cite{margolin2014evaluate} for more details.


\textbf{(iv) Mean Absolute Error~\cite{perazzi2012saliency} (MAE, $M$)} computes the mean absolute error between the predicted COD map $\mathcal{P}_3^c$ and the ground truth $\mathcal{G}^c$ for all image pixels:
\begin{equation}\label{eq_MAE}
    M = \frac{1}{W \times H} \sum_{i=1}^{W}\sum_{j=1}^{H} | \mathcal{P}_{3}^{c}(i, j) - \mathcal{G}^{c}(i, j) |,
\end{equation}
where $W$ and $H$ specify the width and height of $\mathcal{G}^c$.

To conduct a fair comparison, we use the standard evaluation toolbox\footnote{\url{https://github.com/DengPingFan/CODToolbox}} in the COD field to compute all the four metrics of each compared method.

\begin{table*}[t]
\centering
\footnotesize
\renewcommand{\arraystretch}{0.65}
\renewcommand{\tabcolsep}{1.3pt}
\caption{\small Ablation study on CAMO and COD10K datasets. We investigate: 
the baseline (UNet~\cite{ronneberger2015u}+ASPP),
re-calibration process (whether using global-prior (``g-p''), neighbour-prior (``n-p'') or reverse operation (``r-o'') in RRU), 
edge aggregation style (whether employing strategy of Edge Aggregation via Addition (``EAA''), Edge Aggregation via Multiplication (``EAM''), the proposed Selective Edge Aggregation (``SEA'')),
\rev{and multi-context aggregation (whether using Atrous Spatial Pyramid Pooling module~\cite{chen2017deeplab} (``ASPP'') at the head of backbone)}. The best result in each column is highlighted in bold.
}
\label{tab:Ablation}
\begin{tabular}{cc|ccccccc||cccc|cccc}
 \hline
 \toprule
 \rowcolor{mygray}
 \multicolumn{2}{c|}{}
 &\multicolumn{7}{c||}{}
 &\multicolumn{4}{c|}{CAMO~\cite{le2019anabranch}}
 &\multicolumn{4}{c}{COD10K~\cite{fan2020Camouflage}}\\
 \rowcolor{mygray}
   & No.
  & g-p & n-p & r-o
  & EAA & EAM &SEA &ASPP
  &$S_\alpha$      &$E_\phi$     &$F_\beta^w$      &$M$
  &$S_\alpha$      &$E_\phi$     &$F_\beta^w$      &$M$ \\
\hline
\hline
 Base
 &\#1  &               &\CheckmarkBold &             &               &               &  &\CheckmarkBold
 &.739 &.784 &.602 &.103
 &.749 &.825 &.552 &.055 \\
 \hline
 \multirow{3}{*}{RRU}
 &\#2  &\CheckmarkBold &               &\CheckmarkBold &               &               &    &\CheckmarkBold
 &.744 &.791 &.629 &.098
 &.769 &.857 &.617 &.048\\

 &\#3  &\CheckmarkBold &\CheckmarkBold &\CheckmarkBold &               &               &    &\CheckmarkBold
 &.750 &.812 &.639 &.090
 &.774 &.855 &.622 &.044\\

 &\#4  &\CheckmarkBold &\CheckmarkBold &               &               &               &\CheckmarkBold    &\CheckmarkBold
 &.755 &.814 &.658 &.090
 &.777 &.862 &.623 &.046\\
 
 \hline
 \multirow{2}{*}{Edge}
 &\#5  &\CheckmarkBold &\CheckmarkBold &\CheckmarkBold &\CheckmarkBold &               &    &\CheckmarkBold
 &.754 &.812 &.658 &.090
 &.774 &.860 &.619 &.044\\

 &\#6  &\CheckmarkBold &\CheckmarkBold &\CheckmarkBold &               &\CheckmarkBold &    &\CheckmarkBold
 &.752 &.815 &.655 &.092
 &.773 &.854 &.618 &.044\\
 
 \hline
 \multirow{2}{*}{\textbf{Ours}}
 &\rev{\#7}  &\rev{\CheckmarkBold} &\rev{\CheckmarkBold} &\rev{\CheckmarkBold} &               &               &\rev{\CheckmarkBold}    &
 &\rev{.760} &\rev{.814} &\rev{.656} &\rev{\textbf{.087}}
 &\rev{.778} &\rev{.862} &\rev{.627} &\rev{.046}\\
 &\rev{\#8}  &\CheckmarkBold &\CheckmarkBold &\CheckmarkBold &               &               &\CheckmarkBold    &\CheckmarkBold
 &\textbf{.761} &\textbf{.817} &\textbf{.660} &.088
 &\textbf{.780} &\textbf{.867} &\textbf{.629} &\textbf{.044}\\
 \toprule
\end{tabular}
\end{table*}

\subsection{Learning Strategies}\label{sec:learning_strategy}

We implement the model on PyTorch 1.1.0~\cite{paszke2019pytorch} toolbox and all the related experiments are accelerated by an RTX TITAN GPU. We resize the input image and the corresponding ground truth are uniformly resized to 352 $\times$ 352. For data augmentation, we only use multi-scale input images. The weights of ResNet-50~\cite{he2016deep}, pre-trained on ImageNet, are loaded in the training phase, and other layers are initialised randomly. We employ the standard Adam algorithm to optimise the entire network with a learning rate of $1e-4$. The end-to-end training time of the last step takes about 2.5 hours and converges after 30 epochs at all with a batch size of 36. During the inference, we feed the resized image with the shape of 352 $\times$ 352 into~\ourmodel~and obtain results without any post-processing like CRF~\cite{krahenbuhl2011efficient}. The inference speed reaches up to 79.3 FPS.

\vspace{-5pt}
\subsection{Ablation Study}
We conduct thorough ablation studies by decoupling sub-components from the full implementation of~\ourmodel. We set the ResNet-50~\cite{he2016deep} version of ~\ourmodel~as a reference model, and then compare various ablated/modified models to it.

\vspace{0.5mm}
\noindent
\textbf{Effectiveness of RRU.} \
As shown in the pipeline of our network, the reversible re-calibration uint (RRU) at the i$^{th}$ ($i \in \{3,4,5\}$) CNN layer refines its feature map ($E_i$) by considering additional three inputs: the neighbour prior $\mathcal{P}^{c}_{i+1}$, global prior $\mathcal{P}^c_g$, and edge prior $F^e$.
To evaluate the effectiveness of the reversible re-calibration uint (RRU), we conducted an ablation study by simplifying our~\ourmodel~in the following four cases.

\textit{(i) ``baseline'':} we remove all the RRUs at the 3$^{rd}$, 4$^{th}$, and 5$^{th}$ CNN layers from our network while keeping only UNet framework left. Results refer to row \#1 of~\tabref{tab:Ablation}.

\textit{(ii) ``baseline, g-p''}: we add all the proposed RRUs into ``baseline'', but only keep the global prior $\mathcal{P}^c_g$ to obtain a refinement of $E_i$. Results can be found in row \#2.

\textit{(iii) ``baseline, g-p, n-p''}: we keep all the RRUs of our network, but the RRU only takes the global prior $\mathcal{P}^c_g$ and the neighbour prior $\mathcal{P}^{c}_{i+1}$ to refine the feature map. Results are reported in row \#3.
This, in turn, indicates that we remove the edge prior $F^e$ from the RRUs.

\textit{(iv) ``w/o-reverse''}: as shown in \figref{fig:architecture}, RRU first predicts a camouflaged detection map and then reserve it to obtain a reserve attention map indicating non-camouflaged regions, which is then multiplied with the input CNN features $E_i$ for refining it. In order to validate the effectiveness of this reverse operation, we construct a model to remove the reserve attention map from RRU, which means that we directly use the predicted camouflaged detection map to refine $E_i$. Results are shown in row \#4.

Table~\ref{tab:Ablation} reports the four metric results of our full method and the above variants.
Combining the additional three cues (\ie,~global prior, neighbour prior, edge prior) together in the RRU incurs a better COD performance since our method has achieved superior best metric results over the three variants (\ie, \#1, \#2, and \#3) in~\tabref{tab:Ablation}.
Moreover, when RRU with reverse operation (row \rev{\#8}) is compared with the variant without reverse operation (row \#4), our method has better scores than ``\textit{w/o-reverse}'', showing that learning reserve attention in RRU helps our method better identify camouflaged objects.

\vspace{0.5mm}
\noindent
\textbf{Effectiveness of SEA.}
As discussed in Section~\ref{sec:SEA}, the selective edge aggregation (SEA) of our method learns auxiliary edge cues for camouflaged object detection from two low-level features, \ie, $E_1$ and $E_2$.
First, we add edge auxiliary feature into RRU with different feature fusion strategies: one denoted as Edge Aggregation via Addition (``EAA'' in row \#5) is the element-wise addition of $E_1$ and $E_2$ while another denoted as Edge Aggregation via Multiplication (``EAM'' in row \#6) is the element-wise multiplication of $E_1$ and $E_2$. These two variants both perform slightly better than \textit{`baseline+g-prior+n-prior'} variant (row \#3) in terms of $F^w_\beta$ on CAMO, which validate the efficacy of edge information.
Besides, the bottom three rows in Table~\ref{tab:Ablation} report the results of our method and the two models, demonstrating that our method outperforms both ``EAA'' (row \#5) and ``EAM'' (row \#6) in terms of all the four metrics.
It indicates that the SEA can better learn edge information for camouflaged object detection than the simple addition or multiplication fusion strategies.

\vspace{0.5mm}
\noindent
\rev{
\textbf{Effectiveness of ASPP.}
We further ablate the ASPP module (the head of the backbone in \figref{fig:architecture}) to demonstrate the discriminative capability of the proposed network and produce variant \#7 in \tabref{tab:Ablation}.
Compared to our full model (\rev{\#8}), it has slight performance degradation due to the removal of the multi-scale aggregation module.
Besides, as compared with the results shown in \tabref{tab:BenchmarkResults}, we observe that variant \#7 can still achieve the best performance even without the assistance of ASPP.
Thus, the ASPP attributes to the generation of more discriminative multi-context features for the following decoding.
}

\begin{figure*}[t!]
\centering
\includegraphics[width=0.99\textwidth]{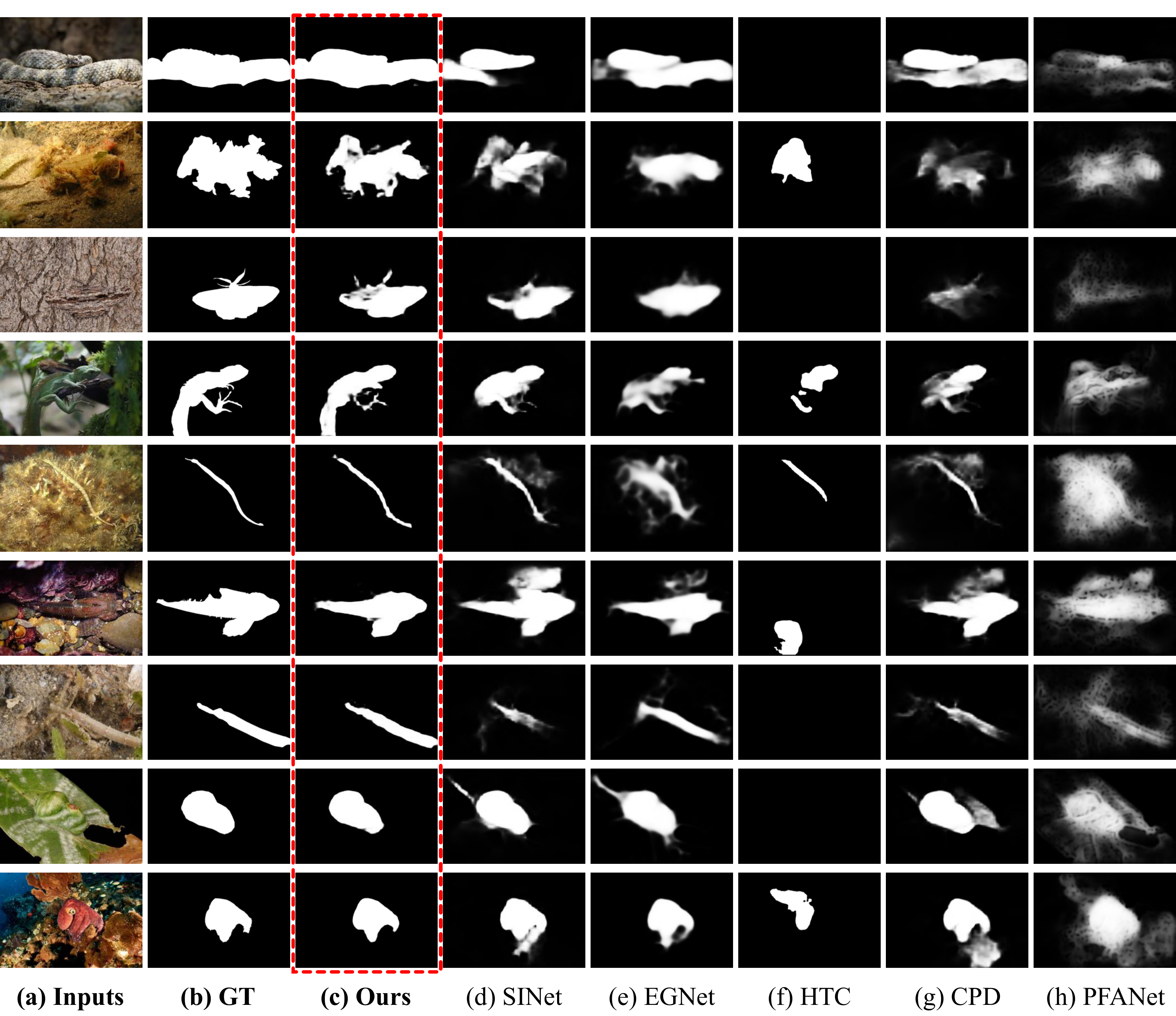}
\vspace{-8pt}
\caption{\small Visual comparison of camouflaged object detection maps produced by different methods. (a) Input images, (b) GT, which stands for the ground truths, (c) camouflaged object detection maps produced by our method, (d) SINet~\cite{fan2020Camouflage}, (e) EGNet~\cite{zhao2019EGNet}, (f) HTC~\cite{chen2019hybrid}, (g) CPD~\cite{wu2019cascaded}, and (h) PFANet~\cite{zhao2019pyramid}.
}
\label{fig:comparison_real_photos_part1}
\end{figure*}

\subsection{Comparisons with SOTA COD Methods}
\textbf{Compared Methods.} \
We compare our method against 13 cutting-edge competitors, including (1) FPN~\cite{lin2017feature}, (2) MaskRC~\cite{he2017mask}, (3) PSPNet~\cite{zhao2017pyramid}, (4) UNet++~\cite{zhou2019unet++}, (5) PiCANet~\cite{liu2018picanet}, (6) MSRC~\cite{huang2019mask}, (7) BASNet~\cite{qin2019basnet}, (8) PFANet~\cite{zhao2019pyramid}, (9) CPD~\cite{wu2019cascaded}, (10) HTC~\cite{chen2019hybrid}, (11) EGNet~\cite{zhao2019EGNet}, (12) A-Net~\cite{le2019anabranch}, and (13) SINet~\cite{fan2020Camouflage}.
As stated in~\cite{fan2020Camouflage}, these baselines are trained from scratch on the COD training set mentioned in~\secref{sec:benchmark_protocols} with the default parameter settings. Hence, we use the standard benchmark results reported in~\cite{fan2020Camouflage}, as shown in~\tabref{tab:BenchmarkResults}.

\noindent
\textbf{Quantitative Comparisons.} \
\tabref{tab:BenchmarkResults} summarises the quantitative results of different COD methods on the three benchmark datasets.
From the results, we can observe that the SINet, as the latest COD detector, has superior metric performance than other competitors on three COD benchmark datasets.
Compared to SINet, our method has higher $S_\alpha$, $E_\phi$, and $F_\beta^w$ scores, as well as smaller $M$ scores, demonstrating that our method can more accurately identify camouflaged object detection.
Specifically, our method has a 4.8\% improvement on the average $E_\phi$ and 6.6\% improvement on the average $F_\beta^w$ for the three benchmark datasets.
The proposed~\ourmodel~has a superior performance over compared 13 SOTA methods on all the three COD datasets, which further demonstrates that our method can more accurately detect camouflaged objects.

\noindent
\textbf{Visual Comparisons.} \
\figref{fig:comparison_real_photos_part1} shows the COD maps produced by our network and the compared models.
The compared methods tend to include many non-camouflaged regions or neglect parts of camouflaged objects in their results.
On the contrary, our method can more accurately detect camouflaged objects, and our results in~\figref{fig:comparison_real_photos_part1} (c) are the closest to the ground truths shown in~\figref{fig:comparison_real_photos_part1} (b).
More visual comparisons on COD datasets can be found in the YouTube~\footnote{\url{https://youtu.be/E8Qsn4fXR7o}}.

\begin{table*}[t!]
	\centering
	\small
	\caption{\small Quantitative results on four super-classes of COD10K~\cite{fan2020Camouflage} datasets, including Amphibians (\textit{Amp.}), Aquatic (\textit{Aqu.}), Flying (\textit{Fly.}), and Terrestrial (\textit{Ter.}). $E_\phi$ denotes mean E-measure~\cite{fan2018enhanced}.}
	\renewcommand{\arraystretch}{0.9}
	\setlength
	\tabcolsep{2.3pt}
		\begin{tabular}{r||ccc|ccc|ccc|ccc}
			\hline
			\toprule
			\rowcolor{mygray}
			&\multicolumn{3}{c|}{COD10K-\textit{Amp.}}
			&\multicolumn{3}{c|}{COD10K-\textit{Aqu.}}
			&\multicolumn{3}{c|}{COD10K-\textit{Fly.}}
			&\multicolumn{3}{c}{COD10K-\textit{Ter.}}\\
			\cline{2-13}
			\rowcolor{mygray}
			Models
			&$S_\alpha\uparrow$      &$E_\phi\uparrow$      &$M\downarrow$
			&$S_\alpha\uparrow$      &$E_\phi\uparrow$      &$M\downarrow$
			&$S_\alpha\uparrow$      &$E_\phi\uparrow$      &$M\downarrow$
			&$S_\alpha\uparrow$      &$E_\phi\uparrow$      &$M\downarrow$\\
			\hline
			\hline
			FPN~\cite{lin2017feature}
			& .744 & .743 & .065 & .684 & .687 & .103 & .726 & .714 & .061 & .668 & .661 & .071 \\
		    MaskRC~\cite{he2017mask}
			& .665 & .782 & .081 & .560 & .719 & .123 & .644 & .765 & .063 & .608 & .747 & .070 \\
			PSPNet~\cite{zhao2017pyramid}
			& .736 & .733 & .072 & .659 & .670 & .111 & .700 & .692 & .067 & .658 & .666 & .074 \\
			UNet++~\cite{zhou2019unet++}
			& .677 & .725 & .079 & .599 & .659 & .121 & .659 & .708 & .068 & .593 & .637 & .081 \\
			PiCANet~\cite{liu2018picanet}
			& .704 & .689 & .086 & .629 & .623 & .120 & .677 & .663 & .076 & .625 & .628 & .084 \\
			MSRC~\cite{huang2019mask}
			& .722 & .784 & .055 & .614 & .685 & .107 & .674 & .742 & .058 & .611 & .671 & .070 \\
			BASNet~\cite{qin2019basnet}
			& .708 & .741 & .087 & .620 & .666 & .134 & .664 & .710 & .086 & .601 & .645 & .109 \\
			PFANet~\cite{zhao2019pyramid}
			& .690 & .661 & .119 & .629 & .614 & .162 & .657 & .632 & .113 & .609 & .600 & .123 \\
			CPD~\cite{wu2019cascaded}
			& .794 & .823 & .051 & .739 & .770 & .082 & .777 & .796 & .046 & .714 & .735 & .058 \\
			HTC~\cite{chen2019hybrid}
			& .606 & .596 & .088 & .507 & .494 & .129 & .582 & .558 & .070 & .530 & .484 & .078 \\
			EGNet~\cite{zhao2019EGNet}
			& .788 & .837 & .048 & .725 & .775 & .080 & .768 & .803 & .044 & .704 & .748 & .054 \\
			SINet~\cite{fan2020Camouflage}
			& \textbf{.827} & .866 & .042 & .758 & .803 & .073 & .798 & .828 & .040 & .743 & .778 & .050 \\
			\hline
			\textbf{\ourmodel}
			& .824 & \textbf{.904} & \textbf{.036} & \textbf{.772} & \textbf{.857} & \textbf{.061} & \textbf{.808} & \textbf{.892} & \textbf{.033} & \textbf{.749} & \textbf{.843} & \textbf{.045} \\
			\hline
			\toprule
	\end{tabular}
	\label{tab:ModelScore_Super_Class}
	\vspace{5pt}
\end{table*}

\noindent
\textbf{Super-class Based Performance on COD10K.} \
For the challenging dataset COD10K~\cite{fan2020Camouflage}, it contains five super-classes: ``Amphibians'', ``Aquatic Animals'', ``Flying Animals'', ``Terrestrial Animals'', and ``Others''.
Due to limited space, we only report the quantitative results of each SOTA methods on four dominant super-classes, as shown in~\tabref{tab:ModelScore_Super_Class}.
We observe that our model significantly surpasses the best method SINet with an impressive gap, \eg, a 5.4\% increase on ``Aquatic Animals'' and 6.4\% on ``Flying Animals'' in terms of $F_{\beta}^w$.

\noindent
\textbf{Analysis of Speed-Accuracy Trade-Off.} \
\figref{fig:inference_speed} shows that our highly efficient model has considerable improvements compared with the existing 13 COD methods. Note that the inference time of models, except for the part of competitors that appeared before 2017 (\ie, FPN~\cite{lin2017feature}, MaskRC~\cite{he2017mask}, and PSPNet~\cite{zhao2017pyramid}), are tested on the same NVIDIA RTX graphics card. Concerning inference speed, the gap between our model (79.3~FPS, $E_\phi=0.867$) and the top-1 model CPD (62~FPS, $E_\phi=0.770$) reach 27.9\%, while improving performance by 9.7\%. Without high complexity modules like dense connections \cite{huang2017densely} or non-local means~\cite{wang2018non}, we only use simple operations (\ie, concatenation, subtraction, and addition) for feature fusion in order for accurate and fast detection. Furthermore, with respect to detection accuracy, our method significantly improves the inference speed with a huge gap compared to the top-1 model SINet (5~FPS, $E_\phi=0.806$), while maintaining a better performance.

\begin{figure*}[t!]
\centering
\includegraphics[width=\textwidth]{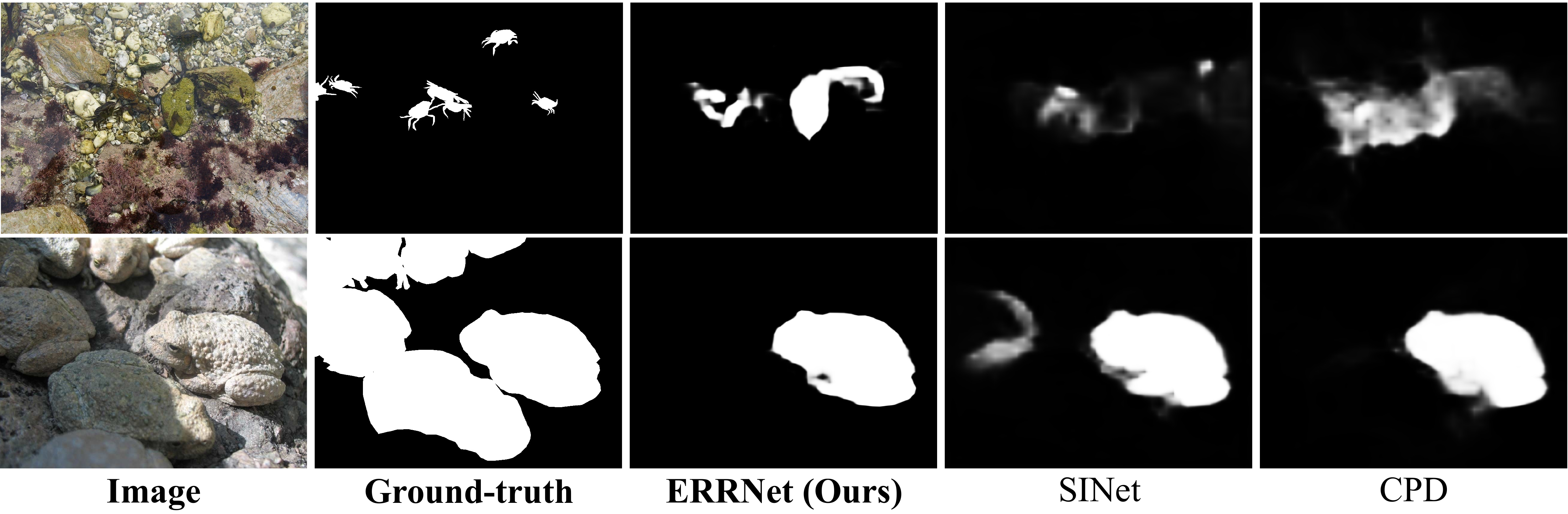}
\caption{\small \rev{Failure cases.}
}
\label{fig:FailureCases}
\end{figure*}

\noindent
\rev{
\textbf{Failure cases in challenging scenes.}
We present failure cases of the proposed \ourmodel~in \figref{fig:FailureCases}. Although satisfactory quantitative performance is achieved, \ourmodel~may fail in challenging scenes containing multiple objects. Specifically, it still difficult to justify when and where multiple camouflaged objects exist, resulting in some false-positive predictions. As can be seen from~\figref{fig:FailureCases}, such cases also easily confuse SOTA approaches, thus deserving further exploration.}

\begin{table}[!t]
  \centering
  \footnotesize
  \renewcommand{\arraystretch}{0.9}
  \setlength\tabcolsep{5.5pt}
  \caption{\small Quantitative results on four widely-used polyp segmentation testing dataset. SFA~\cite{fang2019selective} results are generated using the released code. Note that ``$\dagger$'' represents evaluation scores from~\cite{jha2019resunetplus} and ``N/A'' means the result is not available.
  }\label{tab:polyp_results}
  \begin{tabular}{llr||cccccccc}
  \toprule
  \rowcolor{mygray}
   & &Methods~(Pub./Year)
   & Dice$\uparrow$ & IoU$\uparrow$  &  $F_{\beta}^{w}\uparrow$ & $S_{\alpha}\uparrow$ &$E_\phi\uparrow$ & $M\uparrow$ \\
  \hline
  \multicolumn{9}{c}{$\blacklozenge$~Current SOTA polyp segmentation results on \textbf{\textit{seen dataset}}~$\blacklozenge$} \\
  \hline
  \multirow{7}{*}{\begin{sideways}Kvasir\end{sideways}}
  &\multirow{7}{*}{\begin{sideways}\cite{jha2020kvasir}\end{sideways}}
  &UNet (MICCAI'15)~\cite{ronneberger2015u} & 0.818 & 0.746 & 0.794 & 0.858 & 0.893 & 0.055 \\
  & &UNet++ (TMI'19)~\cite{zhou2019unet++} & 0.821  & 0.743 & 0.808 & 0.862 & 0.910 & 0.048 \\
  & & ResUNet-mod$^\dagger$~\cite{zhang2018road} & 0.791 & N/A & N/A & N/A & N/A & N/A\\
  & & ResUNet++$^\dagger$~\cite{jha2019resunetplus} & 0.813 & 0.793 & N/A & N/A & N/A & N/A \\
  & & SFA (MICCAI'19)~\cite{fang2019selective} & 0.723 & 0.611 & 0.670 & 0.782 & 0.849 & 0.075\\
  & & PraNet (MICCAI'20)~\cite{fan2020pra}
  &0.898 &0.840 & 0.885 & \textbf{0.915} &0.952 & 0.030 \\
  & &\textbf{\ourmodel$^\ddag$~(Ours)}
  &\textbf{0.901} &\textbf{0.840} &\textbf{0.891} &0.911 &\textbf{0.952} &\textbf{0.027} \\

  \hline
  \multirow{8}{*}{\begin{sideways}CVC-612\end{sideways}}
  &\multirow{8}{*}{\begin{sideways}\cite{bernal2015wm}\end{sideways}}
    &UNet (MICCAI'15)~\cite{ronneberger2015u} & 0.823 & 0.755 & 0.811 & 0.889 & 0.954 & 0.019 \\
  & &UNet++ (TMI'19)~\cite{zhou2019unet++} & 0.794  & 0.729 & 0.785 & 0.873 & 0.931 & 0.022\\
  & &ResUNet-mod$^\dagger$~\cite{zhang2018road} & 0.779 & N/A & N/A & N/A & N/A & N/A \\
  & &ResUNet++$^\dagger$~\cite{jha2019resunetplus} & 0.796  & 0.796 & N/A & N/A & N/A & N/A\\
  & &SFA (MICCAI'19)~\cite{fang2019selective} & 0.700 & 0.607 & 0.647 & 0.793 & 0.885 & 0.042\\
  & &PraNet (MICCAI'20)~\cite{fan2020pra} & 0.899 & 0.849 & 0.896 &0.936 & 0.979  &0.009 \\
  & &\textbf{\ourmodel$^\ddag$~(Ours)}
  &\textbf{0.918} &\textbf{0.868} &\textbf{0.917} &\textbf{0.940} &\textbf{0.986} &\textbf{0.009} \\

  \hline
  \multicolumn{9}{c}{$\blacklozenge$~Current SOTA polyp segmentation results on \textbf{\textit{unseen dataset}}~$\blacklozenge$} \\

  \hline
  \multirow{5}{*}{\begin{sideways}ETIS\end{sideways}}
  &\multirow{5}{*}{\begin{sideways}\cite{silva2014toward}\end{sideways}}
  &UNet (MICCAI'15)~\cite{ronneberger2015u}
  & 0.398 & 0.335 & 0.366 & 0.684 & 0.740 & 0.036 \\
  & &UNet++ (TMI'19)~\cite{zhou2019unet++}
  & 0.401 & 0.344 & 0.390 & 0.683 & 0.776 & 0.035 \\
  & &SFA (MICCAI'19)~\cite{fang2019selective}
  & 0.297 & 0.217 & 0.231 & 0.557 & 0.633 & 0.109\\
  & &PraNet (MICCAI'20)~\cite{fan2020pra}
  & 0.628 &0.567 & 0.600 & 0.779  & 0.841 & 0.031 \\
  & &\textbf{\ourmodel$^\ddag$~(Ours)}
  &\textbf{0.691} &\textbf{0.611} &\textbf{0.658} &\textbf{0.822} &\textbf{0.886} &\textbf{0.014} \\

  \hline
  \multirow{5}{*}{\begin{sideways}CVC-T\end{sideways}}
  &\multirow{5}{*}{\begin{sideways}\cite{vazquez2017benchmark}\end{sideways}}
  &UNet (MICCAI'15)~\cite{ronneberger2015u}
  & 0.710 & 0.627 & 0.684 & 0.843 & 0.876 & 0.022\\
  & &UNet++ (TMI'19)~\cite{zhou2019unet++}
  & 0.707 & 0.624 & 0.687 & 0.839 & 0.898 & 0.018\\
  & &SFA (MICCAI'19)~\cite{fang2019selective}
  & 0.467 & 0.329 & 0.341 & 0.640 & 0.817 & 0.065\\
  & &PraNet (MICCAI'20)~\cite{fan2020pra}
  & 0.871 & 0.797 & 0.843 &0.925 & 0.972 & 0.010 \\
  & &\textbf{\ourmodel$^\ddag$~(Ours)}
  &\textbf{0.889}&\textbf{0.813} &\textbf{0.864} &\textbf{0.931} &\textbf{0.978} &\textbf{0.006} \\
  \hline
  \toprule
  \end{tabular}
\end{table}

\begin{table}[!t]
  \centering
  \footnotesize
  \renewcommand{\arraystretch}{0.9}
  \setlength\tabcolsep{6pt}
  \caption{\small Quantitative results on COVID-SemiSeg~\cite{fan2020InfNet} dataset.
  }\label{tab:covid19_Scores}
  \begin{tabular}{l|c||cccc}
  \hline
  \toprule
  \rowcolor{mygray}
   Methods~(Pub./Year) & Backbone
   & Dice$\uparrow$ & $Sen.\uparrow$ & $E_\phi\uparrow$ & $M\downarrow$ \\
  \hline
  UNet~(MICCAI'15)~\cite{ronneberger2015u} & VGGNet-16
  & 0.439 & 0.534 & 0.625 & 0.186\\

  Attention-UNet~(MIDL'18)~\cite{oktay2018attention} & VGGNet-16
  & 0.583 & 0.637 & 0.739 & 0.112\\

  Gated-UNet~(MIDL'18)~\cite{schlemper2019attention} & VGGNet-16
  & 0.623& 0.658 & 0.814 & 0.102\\

  Dense-UNet~(TMI'18)~\cite{li2018h} & DenseNet-161
  & 0.515 & 0.594 & 0.662 & 0.184 \\

  UNet++~(TMI'19)~\cite{zhou2019unet++} & VGGNet-16
  & 0.581  & 0.672 & 0.720 & 0.120 \\

  Inf-Net~(TMI'20)~\cite{fan2020InfNet} & ResNet-50
  & 0.636 & 0.629  & 0.792 & 0.095 \\

  Inf-Net~(TMI'20)~\cite{fan2020InfNet} & Res2Net-50
  &0.682 &0.692 &0.838 &\textbf{0.082} \\

  \hline
  \hline
  \textbf{\ourmodel~(Ours)} & ResNet-50
  &\textbf{0.700} &\textbf{0.751} &\textbf{0.860} &0.084 \\
  \toprule
  \end{tabular}
\end{table}

\begin{figure}[t]
\centering
\includegraphics[width=.90\linewidth]{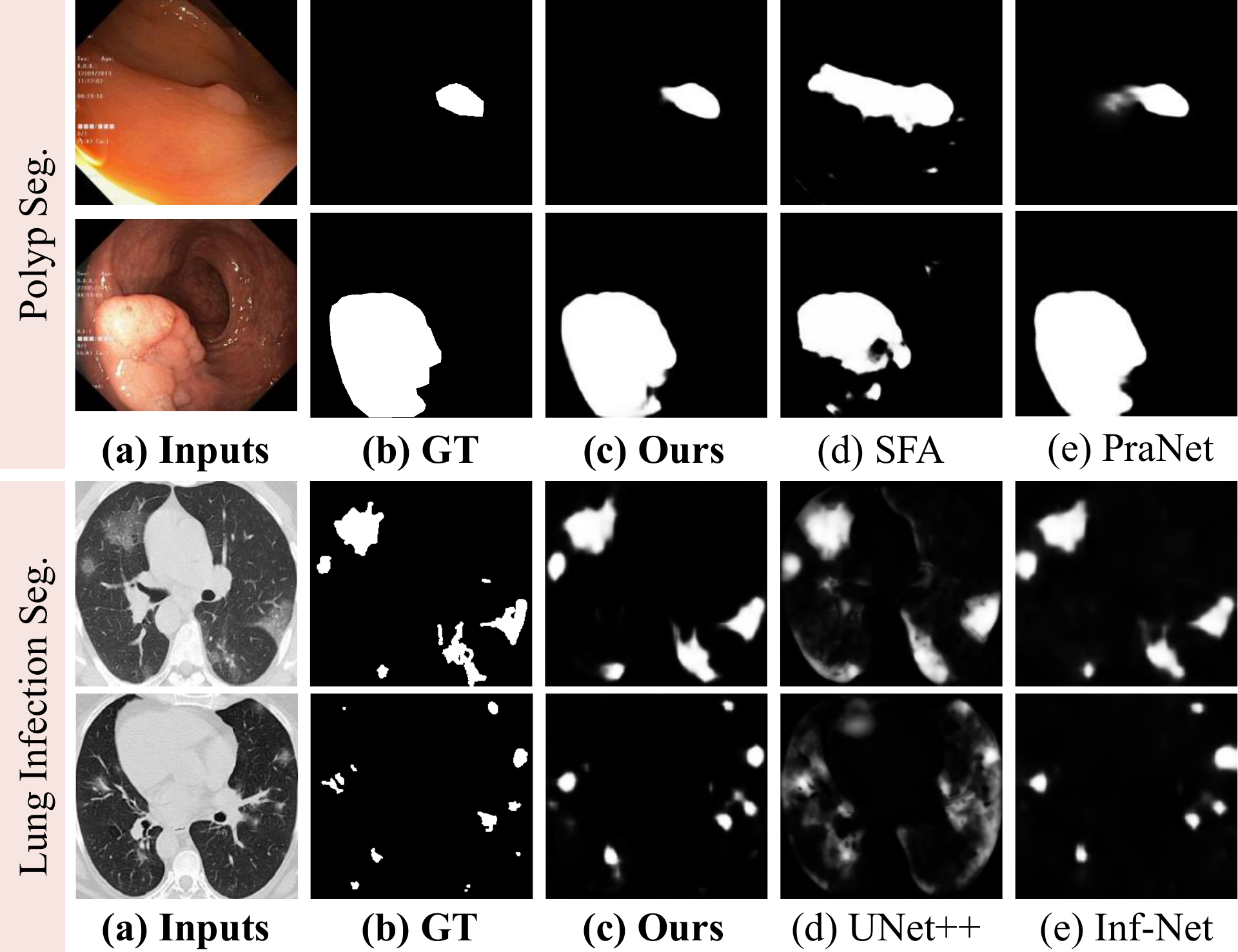}
\caption{\small Qualitative comparison among four cutting-edge methods: SFA~\cite{fang2019selective} (MICCAI'2019), PraNet~\cite{fan2020pra} (MICCAI'2020), UNet++~\cite{zhou2019unet++} (TMI'2019), and Inf-Net~\cite{fan2020InfNet} (TMI'2020); showing results for two distinct medical image segmentation.}
\label{fig:ext_applications}
\vspace{-0.5cm}
\end{figure}

\subsection{Results on Medical Image Segmentation}

Generally, natural camouflaged patterns are abundant in medical clinical diagnosis~\cite{spyropoulou2009decorative,walker2014medical,tanioka2010camouflage} and image processing~\cite{krupinski2010current,mehmood2013prioritization,barata2018survey,celebi2015state}. Therefore, we further evaluate the effectiveness of our~\ourmodel~on the other two binary segmentation tasks in medical imaging in this paper, \ie, Polyp Segmentation and COVID-19 Lung Infection Segmentation.

\noindent
\textbf{Polyp Segmentation.} \
Following the same training/testing protocols in~\cite{jha2019resunetplus,fan2020pra}, we retrain our model on the training set of Kvasir~\cite{jha2020kvasir} and CVC-ClinicDB~\cite{bernal2015wm} dataset from scratch, in which 80\% for training, 10\% for validation, and 10\% for testing.
Moreover, we adopt the same metrics used in~\cite{fan2020pra}, including Dice, IoU, weighted F-measure ($F_{\beta}^{w}$), S-measure ($S_{\alpha}$), E-measure ($E_\phi$), and MAE ($M$).
Testing of the model is conducted on four public polyp segmentation datasets, including \textit{\textbf{seen dataset}} (test part of Kvasir~\cite{jha2020kvasir} and CVC-ClinicDB/CVC-612~\cite{bernal2015wm}) and \textbf{\textit{unseen dataset}} (test part of ETIS~\cite{silva2014toward} and EndoScene/CVC-T~\cite{vazquez2017benchmark}).
We compare our method with 6 SOTA deep-based medical segmentation methods: UNet~\cite{ronneberger2015u}, UNet++~\cite{zhou2019unet++}, ResUNet-mod~\cite{zhang2018road}, ResUNet++~\cite{jha2019resunetplus}, SFA~\cite{fang2019selective} and PraNet~\cite{fan2020pra}.
Since top-1 model PraNet~\cite{fan2020pra} used Res2Net~\cite{gao2019res2net} as the backbone, we retrain~\ourmodel~under the same backbone for a fair comparison, which is marked as ``$\ddagger$'' in \tabref{tab:polyp_results}.
The quantitative comparison results are shown in~\tabref{tab:polyp_results}, and we find that the proposed \ourmodel~also outperforms all the competitors on four datasets consistently. Qualitative results refer to the first two rows of~\figref{fig:ext_applications}.

\noindent
\textbf{Lung Infection Segmentation.} \
We also show how our method can be retrained for COVID-19 lung infection segmentation from CT data.
Follow the same training protocols in the most recent Inf-Net~\cite{fan2020InfNet}, we retrain our model from scratch for fair comparisons.
Meanwhile, we use the same metrics used in~\cite{fan2020InfNet} for quantitative comparisons. They are Dice, Sensitivity ($Sen.$), mean E-measure ($E_\phi$), and MAE ($M$).
Also, we compare our network against six SOTA methods, covering UNet~\cite{ronneberger2015u},
Attention-UNet~\cite{oktay2018attention},
Gated-UNet~\cite{schlemper2019attention},
Dense-UNet~\cite{li2018h},
UNet++~\cite{zhou2019unet++}, and
Inf-Net~\cite{fan2020InfNet}. Details can be found in  \cite{fan2020InfNet}.
~\tabref{tab:covid19_Scores} reports the metric results of different methods, demonstrating that~\ourmodel~has achieved superior scores of all the four metrics overall SOTA approaches.
Especially, compared to the cutting-edge method (\ie, Inf-Net) equipped with a more strong multi-scale backbone (Res2Net~\cite{gao2019res2net}), our~\ourmodel~only with standard ResNet backbone can largely improve the performance of detecting infection regions with a 12\% gain on the $Sen.$ metric.

\section{Conclusion}
In this work, we propose an \textbf{E}dge-based \textbf{R}eversible \textbf{R}e-calibration \textbf{Net}work (\textbf{\ourmodel}) for camouflaged object detection. It consists of selective edge aggregation (SEA) modules and reversible re-calibration units (RRUs) which cooperate closely with NEGS priors (\ie, Neighbour prior, Global prior, Edge prior, and Semantic prior) at both low- and high-level layers. The SEA aggregation strategy is exploited to mine the edge prior and prevent the vanishing problem of weak edges, while the RRU is responsible for re-calibrating the coarse prediction. Our \ourmodel~achieves the 1$^{st}$ place on three COD datasets and also outperforms existing cutting-edge models on five medical image segmentation datasets. Further, we have provided comprehensive ablation studies, showing that \ourmodel~could be a general and robust solution for the COD task. 

However, the camouflage object detection task is still encountering many challenging situations in existing settings. 
\rev{In the future, we plan to explore more low-level patterns such as texture, gradient, and colour information, similar to the object edges in this paper, as auxiliary cues in the deep-based COD scheme.
Besides, we intend to consider the problem of introducing another modality (\eg, depth, or thermal infrared) to further improve detection accuracy like existing RGB-D~\cite{fu2020jl,zhang2021depth,fu2021siamese} and RGB-T~\cite{tu2019rgbt} detection tasks.
}

\section*{Acknowledgements}
This work was supported in part by the NSFC, under No. 62176169, 61703077, the Chengdu Key Research and Development Support Program (2019-YF09-00129-GX), and SCU-Luzhou Municipal People's Government Strategic Cooperation Project (No. 2020CDLZ-10).

\bibliography{CamObjDet_PR}

\end{document}